\documentclass[journal]{IEEEtran}
\usepackage[utf8]{inputenc} % allow utf-8 input
\usepackage[T1]{fontenc}    % use 8-bit T1 fonts
\usepackage{hyperref}       % hyperlinks
\usepackage{url}            % simple URL typesetting
\usepackage{booktabs}       % professional-quality tables
\usepackage{amsfonts}       % blackboard math symbols
\usepackage{nicefrac}       % compact symbols for 1/2, etc.
\usepackage{microtype}      % microtypography
\usepackage{csvsimple}
\usepackage{amsopn}
\usepackage{amsthm}
\usepackage{etex}
\usepackage{graphicx}
\usepackage{endnotes}
\usepackage{hyperref}
\usepackage{epsfig,psfrag}
\usepackage{pst-all}
\usepackage{amssymb,amsfonts,upref,cite,epsf,color,bm}
\usepackage{graphicx}
\usepackage{color}
\usepackage{amsmath}
\usepackage{graphicx}
\usepackage{calc}
\usepackage{booktabs}
\usepackage{tikz}
\usepackage{pgfplots}

\newcommand{\vx}[0]{{\bf x}}
\newcommand{\vv}[0]{{\bf v}}

\newcommand{\vm}[0]{{\bf m}}

\newcommand{\vw}[0]{{\bf w}}

\newcommand{\vy}[0]{{\bf y}}

\newcommand{\bmx}[0]{\begin{bmatrix}}
\newcommand{\emx}[0]{\end{bmatrix}}

\newcommand\defeq{:=}

\newcommand{\featuredim}{n}
\newcommand{\samplesize}{m}
\newcommand{\sampleidx}{i} 
\newcommand{\clusteridx}{c} 
\newcommand{\nrcluster}{k}

\newcommand\dataset{\mathbb{X}}

\newcommand{\emperror}{\mathcal{E}}

\newcommand{\prob}[1]{{\rm P}({#1})}

\DeclareMathOperator*{\argmax}{argmax}
\DeclareMathOperator*{\argmin}{argmin}

\usetikzlibrary{positioning}
\usetikzlibrary{shapes.misc}

\usepackage{algorithm}% http://ctan.org/pkg/algorithms
\usepackage{algpseudocode}% http://ctan.org/pkg/algorithmicx
\floatname{algorithm}{Algorithm}
\algnewcommand\algorithmicinput{\textbf{Input:}}
\algnewcommand\INPUT{\item[\algorithmicinput]}
\algnewcommand\algorithmicoutput{\textbf{Output:}}
\algnewcommand\OUTPUT{\item[\algorithmicoutput]}

\newcommand{\featurelen}{\featuredim}

\newcommand{\cluster}{\mathcal{C}}
\newcommand{\nrclusters}{k} 
\newcommand{\clusterassgt}{y}

\newcommand{\boundellipse}[3]% center, xdim, ydim
{(#1) ellipse (#2 and #3)
}
\def\biglen{20cm} % playing role of infinity (should be < .25\maxdimen)
\tikzset{
  half plane/.style={ to path={
       ($(\tikztostart)!.5!(\tikztotarget)!#1!(\tikztotarget)!\biglen!90:(\tikztotarget)$)
    -- ($(\tikztostart)!.5!(\tikztotarget)!#1!(\tikztotarget)!\biglen!-90:(\tikztotarget)$)
    -- ([turn]0,2*\biglen) -- ([turn]0,2*\biglen) -- cycle}},
  half plane/.default={1pt}
}

\tikzset{cross/.style={cross out, draw=black, minimum size=2*(#1-\pgflinewidth), inner sep=0pt, outer sep=0pt},
cross/.default={3pt}}

\usepackage{setspace}

\title{Basic Principles of Clustering Methods}

\author{Alexander Jung and Ivan Baranov  \\[3mm]
Department of Computer Science \\ 
Aalto University, Finland
}
\begin{document}
%\thanks{\hspace*{-5mm}The work of ??? was supported by ???.} 
	\maketitle
\begin{abstract}
Clustering methods group a set of data points into a few coherent groups or clusters 
of similar data points. As an example, consider clustering pixels in an image (or video) if they 
belong to the same object. Different clustering methods are obtained by using different notions 
of similarity and different representations of data points. 

\end{abstract}

\section{Introduction}
\label{sec_intro}

We consider data points which can be represented by Euclidean vectors $\vx=\big(x_{1},\dots,x_{\featurelen}\big)^{T} \in \mathbb{R}^{\featurelen}$. 
For digital images and videos such a representation can be obtained by stacking the red, green and blue intensities of pixels into a vector \cite{ImageNet}. 
Data points obtained from sound signals can be transformed into vectors by sampling the signal with a certain sampling frequency. 
Music recordings are typically sampled at a rate of $44 \cdot 10^{3} {\rm Hz}$ \cite{OppenheimSchaferBuck1998}. 
Wireless communication systems face data points in the form of time signals that represent individual transmit symbols. 
These time signals have to sampled at rates proportional to 
the radio bandwidth used in the wireless communication system \cite{FundWireless}. 
Natural language processing deals with data points representing text documents, 
ranging from short text messages to entire document collection \cite{Goodfellow-et-al-2016,Mikolov2013}. 

\begin{figure}[htbp]
	\centering
	\includegraphics[width=8cm]{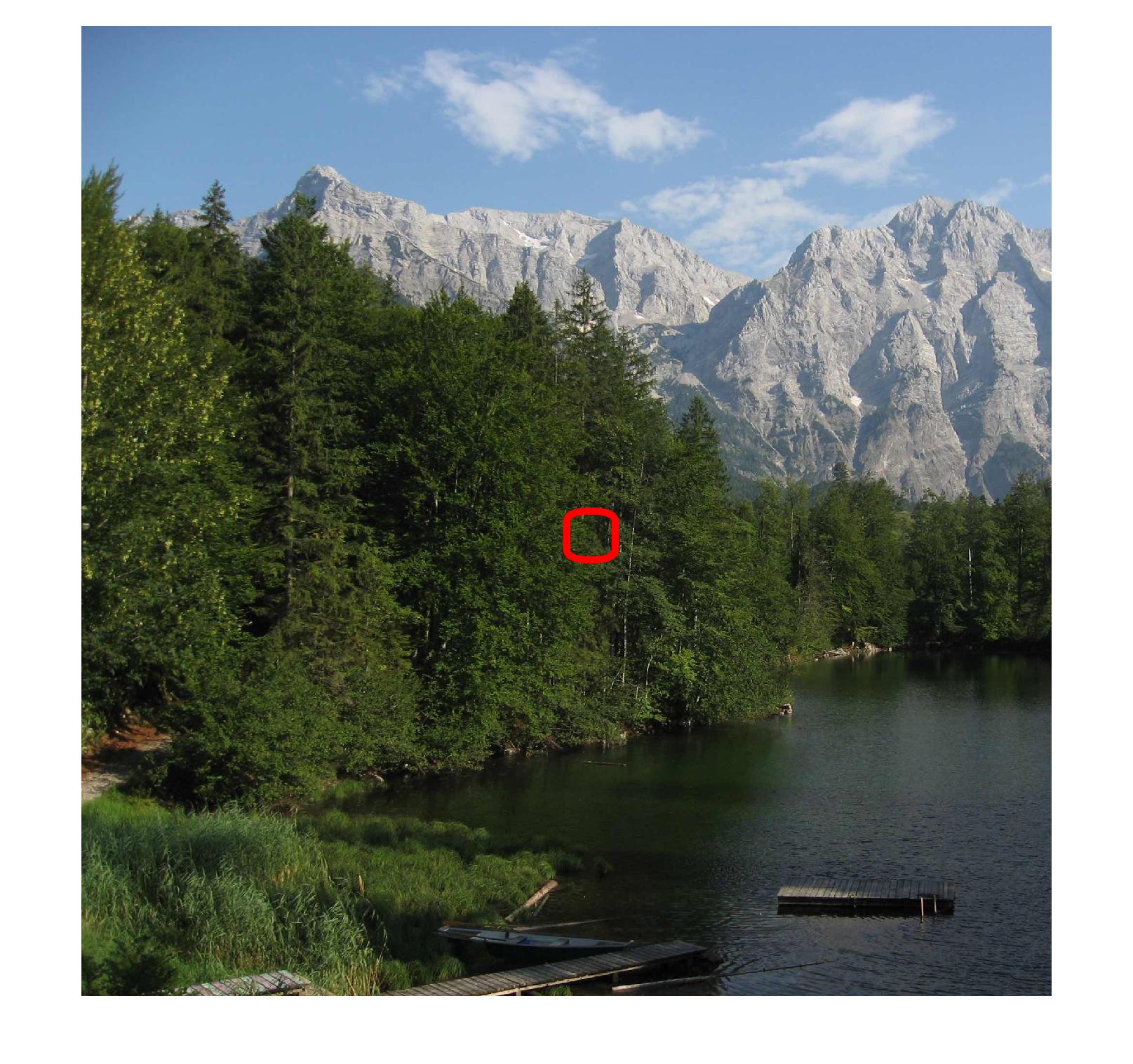}
	\caption{We consider data points obtained from small rectangular regions (``patches'') of a snapshot. One such patch is indicated by a red box.}
	\label{fig:bergseepatch}
\end{figure}

In what follows, we assume that data points are given (represented) in the form of feature vectors $\vx \in \mathbb{R}^{\featurelen}$. 
To have a concrete example in mind, we consider data points representing patches (rectangular regions) of the snapshot depicted in Figure \ref{fig:bergseepatch}. Each 
patch is characterized by the feature vector $\vx = (x_{\rm red},x_{\rm green},x_{\rm blue})$ whose entries are the average 
redness $x_{\rm red}$, greenness $x_{\rm blue}$ and greenness $x_{\rm green}$ of all the pixels belonging to this patch. 
The entire dataset $\mathbb{X} = \{ \mathbf{x}^{(\sampleidx)} \}_{\sampleidx=1}^{\samplesize}$ is given by all the feature 
vectors for the patches. 

\begin{figure}[htbp]
\begin{center}
\begin{tikzpicture}
\tikzstyle{place1}=[circle,draw=blue!50,line width=1.2pt,
                   inner sep=0pt,minimum size=2mm]
 \tikzset{x=5cm,y=5cm,every path/.style={>=latex},node style/.style={circle,draw}}
    \csvreader[ head to column names,%
                late after head=\xdef\aold{\red}\xdef\bold{\green}, filter={\value{csvrow}<20},,%
                after line=\xdef\aold{\red}\xdef\bold{\green}]%
               % {MSEoverBoundary.csv}{}
                {Patches.csv}{}
                {%draw [line width=0.0mm] (\aold, \bold) (\xa,\xb) node {\large $\circ$};
                 \node[place1] at (\red,\blue) (x) {} ;}
              %  \draw [line width=0.0mm] (\hataold, \hatbold) (\hatxa,\hatxb) node {\large $\times$};
%        \csvreader[ head to column names,%
%                late after head=\xdef\aold{\xa}\xdef\bold{\xb},,%
%                after line=\xdef\aold{\xa}\xdef\bold{\xb}]%
%               % {MSEoverBoundary.csv}{}
%                {PrincipalDirection.csv}{}
%                {\draw [line width=0.0mm] (\aold, \bold) -- (\xa,\xb);
%    }
%    
        \draw[->] (-0.1,0) -- (1.1,0);
          \draw[->] (0,-0.1,0) -- (0,1.1);
     %  \node[ align=left,   below] at (0.2,0.2)   {{$\circ$ LP \eqref{equ_LP_problem}}\\{$\times$ Alg.\ \ref{alg_sparse_label_propagation_centralized}}} ; 
 
     \node [anchor=north] at (1.1,-0) {redness};
        \node [anchor=east] at (0,1.1) {blueness};
   %   \draw[->] (0,0) -- (0,4.2);
     
%      %%%% y-axis ticks
%      
%      \node [anchor=south] at (0,2.2) {NMSE $\varepsilon$};
%            \foreach \label/\labelval in {0/$0$,0.5/$0.5$,1/$1$,1.5/$1.5$,2/$2$}
%        { 
%          \draw (1pt,\label) -- (-1pt,\label) node[left] {\large \labelval};
%        }
%        
%        %%% x-axis ticks
%              \foreach \label/\labelval in {0/$0$,0.5/$0.5$,1/$1$,1.5/$1.5$,2/$2$,2.5/$2.5$,3/$3$}
%        { 
%          \draw (\label,1pt) -- (\label,-2pt) node[below] {\large \labelval};
%        }
        
\end{tikzpicture}
        \vspace*{-4mm}
\end{center}
  \caption{Data points representing patches of the image in Figure \ref{fig:bergseepatch}. Each data point is depicted as a circle whose location is given by the average redness and blueness of the pixels in the corresponding image patch. 
  %(normalized average flow between 
  %labeled nodes and the cluster boundaries) of the clusters.
   }
 % increasing variance $\sigma_{e}^{2}$ of the random edge weights of graph $\graph_{1}$ (see \eqref{equ_def_random_weight}).}
  \label{fig_patches_scatter}
  \vspace*{-3mm}
\end{figure}
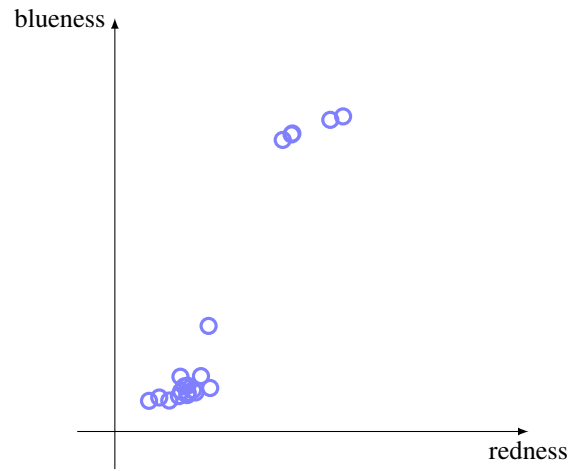

Clustering methods partition a dataset $\mathbb{X} = \{ \mathbf{x}^{(\sampleidx)} \}_{\sampleidx=1}^{\samplesize}$, consisting of $\samplesize$ 
data points represented by the (feature) vectors $\mathbf{x}^{(\sampleidx)} \in \mathbb{R}^{\featurelen}$, into a small number of groups 
or "clusters" $\cluster^{(1)},\ldots,\cluster^{(\nrclusters)}$. 
Each cluster $\cluster^{(\clusteridx)}$ represents a subset of data points which are more similar to each other than to 
data points in another cluster. The precise meaning of calling two data points "similar" depends on the application at 
hand. There are different notions of similarity that lead to different clustering methods. 

Clustering methods do not require labeled data and can be applied to data points which are characterized solely by features 
$\mathbf{x}^{(\sampleidx)}$. Therefore, clustering methods are referred to as \emph{unsupervised machine learning} methods. 
However, clustering methods are often used in combination with (as a preprocessing step for) supervised learning 
methods such as regression or classification. 

There are two main flavours of clustering methods: 
\begin{itemize} 
\item {\bf hard clustering} methods assign each data point to exactly one cluster. 
\item {\bf soft clustering} methods assign each data point to several different clusters 
but with varying degrees of belonging.
\end{itemize}
Hard clustering can be interpreted as a special case of soft clustering 
with the degrees of belonging enforced to be either zero (``no belonging'') or one (``full belonging'').

In Section \ref{sec_hard_clustering} we discuss one popular method for hard clustering, the k-means algorithm. In Section \ref{sec_soft_clustering}, 
we discuss one popular method for soft clustering, which is based on a probabilistic Gaussian mixture model (GMM). 
The clustering methods in Section \ref{sec_hard_clustering} and Section \ref{sec_soft_clustering} use a notion of similarity that is tied to the 
Euclidean geometry of $\mathbb{R}^{\featuredim}$. In some applications, it is more useful to use a different 
notion of similarity. Section \ref{sec_density_clustering} discusses one example of a clustering method which uses 
a non-Euclidean notion of similarity.

%{\bf Notation.} %We denote matrices and vectors using boldface upper and lower case letters. 
%The identity matrix of size $d\!\times\!d$ is denoted $\mathbf{I}_{d}$. 

%\begin{figure}[htbp]
%\begin{center}
%\begin{tikzpicture}[auto,scale=0.8]
%\draw [thick] (5,2.5) rectangle ++(0.1cm,0.1cm) node[anchor=west] {\hspace*{0mm}$\vx^{(3)}$};
%\draw [thick] (4,2) rectangle ++(0.1cm,0.1cm) node[anchor=west] {\hspace*{0mm}$\vx^{(4)}$};
%\draw [thick] (5,1) rectangle ++(0.1cm,0.1cm) node[anchor=west,above] {\hspace*{0mm}$\vx^{(2)}$};
%\draw [thick] (1,5) rectangle ++(0.1cm,0.1cm) node[anchor=west,above] {\hspace*{0mm}$\vx^{(1)}$};
%\draw [thick] (1,3.5) rectangle ++(0.1cm,0.1cm) node[anchor=west,above] {\hspace*{0mm}$\vx^{(5)}$};
%\draw [thick] (1,2.5) rectangle ++(0.1cm,0.1cm) node[anchor=west,above] {\hspace*{0mm}$\vx^{(6)}$};
%\draw [thick] (2,4) rectangle ++(0.1cm,0.1cm) node[anchor=west,above] {\hspace*{0mm}$\vx^{(7)}$};
%\draw[->] (-0.5,0) -- (6.5,0) node[right] {$x_{\rm g}$};
%\draw[->] (0,-0.5) -- (0,6.5) node[above] {$x_{\rm r}$};
%%\foreach \y/\ytext in { 1/1,2/2,3/3,4/4,5/5} \draw[shift={(0,\y)}] (2pt,0pt) -- (-2pt,0pt) node[left] {$\ytext/5$};  
%%\foreach \x/\xtext in{ 1/1,2/2,3/3,4/4,5/5}\draw[shift={(\x,0)}] (0pt,2pt) -- (0pt,-2pt) node[below] {$\xtext/5$};  
%\end{tikzpicture}
%\end{center}
%\caption{A scatterplot obtained from the features $\vx^{(\sampleidx)}=(x^{(\sampleidx)}_{\rm r},x^{(\sampleidx)}_{\rm g})^{T}$, given by the 
%redness $x_{\rm r}^{(\sampleidx)}$ and greenness $x_{\rm g}^{(\sampleidx)}$, of some snapshots. 
%} 
%\label{fig_scatterplot_clustering}
%\end{figure}

\section{Hard Clustering}
\label{sec_hard_clustering}

Hard clustering methods partition a dataset $\mathbb{X}=\{\mathbf{x}^{(1}),\ldots,\mathbf{x}^{(\samplesize)}\}$ into $\nrclusters$ non-overlapping 
clusters $\cluster^{(1)},\ldots,\cluster^{(\nrclusters)}$. Each data point $\mathbf{x}^{(\sampleidx)}$ is assigned to precisely one cluster whose 
index we denote by $y^{(\sampleidx)} \in \{1,\ldots,\nrclusters\}$. %the index of the cluster to which the $i$th data point $\mathbf{x}^{(\sampleidx)}$ belongs to. 

The formal setup of hard clustering is somewhat similar to that of classification methods. Indeed, we can interpret the cluster 
index $y^{(i)}$ as the label (quantity of interest) associated with the $i$ th data point. However, in contrast to classification 
problems, clustering does not require any labeled data points. 

Clustering methods do not require knowledge of the cluster assignment for any data point. Instead, clustering 
methods learn a reasonable cluster assignment for a data point based on the intrinsic geometry of the entire dataset 
$\mathbb{X}$. 

\subsection{$k$-means}
Maybe the most basic (and popular) method for hard clustering is the $k$-means algorithm. 
This algorithm represents each (non-empty) cluster $\cluster^{(\clusteridx)} \subseteq \mathbb{X}$ 
by the cluster mean 
\begin{equation*}
\mathbf{m}^{(\clusteridx)} = (1/|\cluster^{(\clusteridx)} |) \sum_{\mathbf{x}^{(i)} \in\cluster^{(\clusteridx)} } \mathbf{x}^{(i)}. 
\end{equation*}

If we would know the cluster means $\mathbf{m}^{(\clusteridx)}$ for each cluster, we could assign each data point 
$\mathbf{x}^{(\sampleidx)}$ to the cluster with index $\clusterassgt^{(\sampleidx)}$ whose mean $\mathbf{m}^{(\clusterassgt^{(\sampleidx)})}$ 
is closest in Euclidean norm,  
\begin{equation}
\nonumber
\| \mathbf{x}^{(\sampleidx)} - \mathbf{m}^{(\clusterassgt^{(\sampleidx)})} \| =\min\limits_{\clusteridx \in \{1,\ldots,\nrclusters\}}\| \mathbf{x}^{(\sampleidx)} - \mathbf{m}^{(\clusteridx)} \|. 
\end{equation} 
However, in order to determine the cluster means $\mathbf{m}^{(\clusteridx)}$, we would have needed the cluster 
assignments $y^{(i)}$ already in the first place. This instance of an ``egg-chicken dilemma'' (see \url{https://en.wikipedia.org/wiki/Chicken_or_the_egg}) 
is resolved within the $k$-means algorithm by iterating the cluster means and cluster assignment update. We have 
summarized the $k$-means algorithm in Algorithm \ref{alg_k_means}. 

\begin{algorithm}[h]
\caption{$k$-Means Clustering}{}
\begin{algorithmic}[1]
\renewcommand{\algorithmicrequire}{\textbf{Input:}}
\renewcommand{\algorithmicensure}{\textbf{Output:}}
\Require data points $\mathbf{x}^{(1)},\ldots,\mathbf{x}^{(\samplesize)}$, number $\nrclusters$ of clusters, 
initial cluster means $\mathbf{m}^{(\clusteridx)}$ for $\clusteridx=1,\ldots,\nrclusters$. 
%\vspace*{3mm}
%\Statex\hspace{-6mm}{\bf Initialize:} $r\!\defeq\!0$
%\vspace*{1mm}
\Statex \hspace{-7.1mm}\textbf{Iterate}
\Repeat
\vspace*{2mm}
\State update cluster assignments
\begin{equation} 
\label{equ_cluster_assign_update}
y^{(\sampleidx)} \defeq \argmin\limits_{\clusteridx \in \{1,\ldots,\nrclusters\}}\| \mathbf{x}^{(\sampleidx)} - \mathbf{m}^{(\clusteridx)} \| \mbox{ for all } \sampleidx=1,\ldots,\samplesize
\end{equation} 
\State update cluster means  
\begin{equation}  
\label{equ_cluster_mean_update}
   \hspace*{-5mm} \mathbf{m}^{(\clusteridx)}\!\defeq\!\frac{1}{\mid\{i:\!\clusterassgt^{(\sampleidx)}\!=\!\clusteridx\}\mid} \sum_{\sampleidx: \clusterassgt^{(\sampleidx)}\!=\!\clusteridx}\mathbf{x}^{(\sampleidx)}     \mbox{ for all } \clusteridx=1,\ldots,\nrclusters
 \end{equation}
%\vspace*{2mm}
%\State $r \defeq r\!+\!1$    
\vspace*{2mm}
\Until{stopping criterion is satisfied}
\vspace*{1mm}
\Ensure cluster assignments $\clusterassgt^{(\sampleidx)}$ for all $\sampleidx=1,\ldots,\samplesize$
\end{algorithmic}
\label{alg_k_means}
\vspace*{-1mm}
\end{algorithm}

In order to implement Algorithm \ref{alg_k_means}, we need to fix three issues that are somewhat hidden 
in its formulation. First, we need to specify how to break ties in the case when, for a particular data point $\vx^{(\sampleidx)}$, 
there are several cluster indices $\clusteridx\!\in\!\{1,\ldots,\nrcluster\}$ achieving the minimum in \eqref{equ_cluster_assign_update}. 

Second, we need to take care of a situation when after a cluster assignment update \eqref{equ_cluster_assign_update}, 
some cluster $\cluster^{(\clusteridx')}$ has no data points associated with it, 
\begin{equation} 
|\{ \sampleidx: \clusterassgt^{(\sampleidx)} = \clusteridx' \}|=0.
\end{equation}  
In this case, the cluster means update \eqref{equ_cluster_mean_update} would not be well defined for the cluster $\cluster^{(\clusteridx')}$. 
Finally, we need to specify a stopping criterion (``checking convergence''). 

Algorithm \ref{alg:kmeansimpl} is obtained from Algorithm \ref{alg_k_means} by fixing those three issues 
in a particular way \cite{Gray1980}. To this end, it maintains the cluster activity indicators $b^{(\clusteridx)} \in \{0,1\}$, for $\clusteridx=1,\ldots,\nrclusters$. 
Whenever the assignment update \eqref{equ_cluster_assign_update2} results in a cluster $\cluster^{(\clusteridx)}$ 
with no data points assigned to it, the corresponding indicator is set to zero, $b^{(\clusteridx)}=0$. On the other 
hand, if at least one data point is assigned to cluster $\cluster^{(\clusteridx)}$, we
set $b^{(\clusteridx)}=1$. Using these additional variables allows to execute the cluster mean update step \eqref{equ_cluster_mean_update2} only for 
clusters with at least one data point $\vx^{(\sampleidx)}$ assigned to it. The cluster assignment update step \eqref{equ_cluster_assign_update2} 
breaks ties in \eqref{equ_cluster_assign_update} by preferring (other things being equal) lower cluster indices.  
\begin{algorithm}[htbp]
\caption{$k$-Means ``Fixed Point Algorithm'' in \cite{Gray1980}}\label{alg:kmeansimpl}
\begin{algorithmic}[1]
\renewcommand{\algorithmicrequire}{\textbf{Input:}}
\renewcommand{\algorithmicensure}{\textbf{Output:}}
\Require    data points $\mathbf{x}^{(1)},\ldots,\mathbf{x}^{(\samplesize)}$, number $\nrclusters$ of clusters, 
initial cluster means $\mathbf{m}^{(\clusteridx)}$ for $\clusteridx=1,\ldots,\nrclusters$. 
\Repeat 
\State update cluster assignments
\begin{equation} 
\label{equ_cluster_assign_update2}
\hspace*{-2mm}\clusterassgt^{(\sampleidx)}\!\defeq\!\min \{  \argmin\limits_{\clusteridx' \in \{1,\ldots,\nrcluster\}} \| \vx^{(\sampleidx)} - \vm^{(\clusteridx')} \| \} \mbox{ for all }\sampleidx\!=\!1,\ldots,\samplesize  
\end{equation}
\State update activity indicators 
\begin{equation} 
b^{(\clusteridx)} \defeq \begin{cases} 1 & \mbox{ if } |\{i: \clusterassgt^{(\sampleidx)}= \clusteridx\}| > 0 \\ 0 & \mbox{ else.} \end{cases} \mbox{ for all } \clusteridx\!=\!1,\ldots,\nrcluster,
\end{equation}  
\State update cluster means 
\begin{align}
\label{equ_cluster_mean_update2} 
\vm^{(\clusteridx)} & \defeq \frac{1}{|\{i: \clusterassgt^{(\sampleidx)}= \clusteridx\}|}  \sum_{\sampleidx: \clusterassgt^{(\sampleidx)} = \clusteridx} \vx^{(\sampleidx)}  \nonumber  \\
&  \mbox{ for all } \clusteridx\!=\!1,\ldots,\nrcluster \mbox{ with } b^{(\clusteridx)}=1
\end{align} 
%\State $r \defeq r +1$  (increment iteration counter)
%\State $E^{(r)} = \emperror \big( \{\vm^{(\clusteridx)}\}_{\clusteridx=1}^{\nrcluster},\{y^{(\sampleidx)}\}_{\sampleidx=1}^{\samplesize} \mid \dataset \big)$  (see %\eqref{equ_def_emp_risk_kmeans})
\Until cluster assignments $\clusterassgt^{(\sampleidx)}$ do not change
\Ensure cluster assignments $\clusterassgt^{(\sampleidx)}$
\end{algorithmic}
\end{algorithm}

As verified in \cite[Appendix C]{Gray1980}, Algorithm \ref{alg:kmeansimpl} is guaranteed to terminate 
within a finite number of iterations. In other words, after a finite number of cluster mean and assignment updates, 
Algorithm \ref{alg:kmeansimpl} finds a cluster assignments $\{ \clusterassgt^{(\sampleidx)} \}_{i=1}^{\samplesize}$ 
which is unaltered by any additional iteration of cluster mean and assignment update.  

%We illustrate the operation of Algorithm \ref{alg:kmeansimpl} in Figure \ref{fig:first_iter_kmeans}. Each column 
%corresponds to one iteration of Algorithm \ref{alg:kmeansimpl}. The upper picture in each column depicts the 
%update of cluster means while the lower picture shows the update of the cluster assignments during each iteration. 
%
%While Algorithm \ref{alg:kmeansimpl} is guaranteed to terminate after a finite number of iterations, the delivered 
%cluster assignments and cluster means might only be (approximations) of local minima of the clustering error 
%\eqref{equ_def_emp_risk_kmeans} (see Figure \ref{fig_emp_risk_k_means}). In order to escape local minima, 
%it is useful to run Algorithm \ref{alg:kmeansimpl} several times, using different initializations for the cluster means, 
%and picking the cluster assignments $\{ y^{(\sampleidx)} \}_{\sampleidx=1}^{\samplesize}$ with smallest clustering 
%error \eqref{equ_def_emp_risk_kmeans}. 

\subsection{Initialization}

\begin{figure}[htbp]
\begin{center}
\begin{tikzpicture}
\tikzstyle{place1}=[circle,draw=blue!50,line width=1.2pt,
                   inner sep=0pt,minimum size=2mm]
 \tikzset{x=1cm,y=1cm,every path/.style={>=latex},node style/.style={circle,draw}}
    \csvreader[ head to column names,%
                late after head=\xdef\aold{\xa}\xdef\bold{\xb}\xdef\hataold{\hatxa}\xdef\hatbold{\hatxb},,%
                after line=\xdef\aold{\xa}\xdef\bold{\xb}\xdef\hataold{\hatxa}\xdef\hatbold{\hatxb}]%
               % {MSEoverBoundary.csv}{}
                {Dataset.csv}{}
                {%draw [line width=0.0mm] (\aold, \bold) (\xa,\xb) node {\large $\circ$};
                 \node[place1] at (\xa,\xb) (x) {} ;
                \draw [line width=0.0mm] (\hataold, \hatbold) (\hatxa,\hatxb) node {\large $\times$};
    }
    
        \csvreader[ head to column names,%
                late after head=\xdef\aold{\xa}\xdef\bold{\xb},,%
                after line=\xdef\aold{\xa}\xdef\bold{\xb}]%
               % {MSEoverBoundary.csv}{}
                {PrincipalDirection.csv}{}
                {\draw [line width=0.0mm] (\aold, \bold) -- (\xa,\xb);
    }

     %     \draw[->] (0,0) -- (4.2,0);
     %  \node[ align=left,   below] at (0.2,0.2)   {{$\circ$ LP \eqref{equ_LP_problem}}\\{$\times$ Alg.\ \ref{alg_sparse_label_propagation_centralized}}} ; 
 
    %  \node [] at (1.5,-0.5) {\centering cluster connectivity $\bar{\rho}$};
   %   \draw[->] (0,0) -- (0,4.2);
      
%      %%%% y-axis ticks
%      
%      \node [anchor=south] at (0,2.2) {NMSE $\varepsilon$};
%            \foreach \label/\labelval in {0/$0$,0.5/$0.5$,1/$1$,1.5/$1.5$,2/$2$}
%        { 
%          \draw (1pt,\label) -- (-1pt,\label) node[left] {\large \labelval};
%        }
%        
%        %%% x-axis ticks
%              \foreach \label/\labelval in {0/$0$,0.5/$0.5$,1/$1$,1.5/$1.5$,2/$2$,2.5/$2.5$,3/$3$}
%        { 
%          \draw (\label,1pt) -- (\label,-2pt) node[below] {\large \labelval};
%        }
        
\end{tikzpicture}
        \vspace*{-4mm}
\end{center}
  \caption{A dataset along with the subspace spanned by the principal direction. Data points 
  are depicted by circles and the corresponding principal components by crosses. 
  %(normalized average flow between 
  %labeled nodes and the cluster boundaries) of the clusters.
   }
 % increasing variance $\sigma_{e}^{2}$ of the random edge weights of graph $\graph_{1}$ (see \eqref{equ_def_random_weight}).}
  \label{fig_NMSEconnect}
  \vspace*{-3mm}
\end{figure}
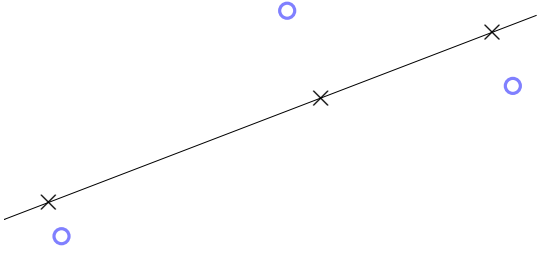

Algorithm \ref{alg_k_means} requires as input some initial choices for the cluster means $\vm^{(\clusteridx)}$, for $\clusteridx=1,\ldots,\nrclusters$. 
Different strategies are used for choosing the initial cluster means. 
%One approach is to model the dataset as realizations of a random vector  i.i.d.\ realizations of a 
%random vector $\vm$ whose distribution is matched to the dataset 
%$\dataset = \{ \vx^{(\sampleidx)} \}_{\sampleidx=1}^{\samplesize}$, e.g., $\vm \sim \mathcal{N}( \hat{\vm}, \widehat{\mathbf{C}})$ 
%with sample mean 
%\begin{equation} 
%\label{equ_sample_means} 
%\widehat{\vm} = (1/\samplesize) \sum_{i=1}^{\samplesize} \vx^{(\sampleidx)} 
%\end{equation}
%and the sample covariance 
%\begin{equation} 
%\label{equ_sample_cov}
%\widehat{\mathbf{C}} = (1/\samplesize) \sum_{i=1}^{\samplesize} (\vx^{(\sampleidx)}\!-\!\hat{\vm}) (\vx^{(\sampleidx)}\!-\!\hat{\vm})^{T}.
%\end{equation}  
 We can initialize the cluster means $\vm^{(\clusteridx)}$ by randomly selecting $k$ different data points 
 $\vx^{(\sampleidx)}$ out of $\dataset$. Another option for initializing the cluster means is  based on the 
 principal components of the data points in $\dataset$. This principal component is found by projecting onto the 
principal direction 
\begin{equation} 
\mathbf{v} = \argmax_{\vw \in \mathbb{R}^{\featurelen}} (1/\samplesize) \sum_{\sampleidx=1}^{\samplesize} \big( \vw^{T} \mathbf{x}^{(\sampleidx)} \big)^{2}.  
\end{equation} 
%Here, we used the sample covariance matrix $\widehat{\mathbf{C}}$ (see \eqref{equ_sample_cov}). 
The principal direction $\mathbf{v}$ spans the one-dimensional subspace $\{ t \mathbf{v}: t \in \mathbb{R} \}$ which is 
the best approximation of $\dataset$ among all one-dimensional subspaces. The cluster means might then be chosen
by evenly partitioning the value range of the principal components $\vv^{T} \vx^{(\sampleidx)}$ for $\sampleidx=1,\ldots,\samplesize$. 
Figure \ref{fig_NMSEconnect} depicts a dataset whose data points are indicated by circles and 
the subspace spanned by the principal direction as a line.

%
%* __Input__: data points $\mathbf{x}^{(i)} \in \mathbb{R}^{n}$, for $i=1,\ldots,m$ and number $k$ of clusters
%
%
%* __Initialization__: choose initial cluster means $\mathbf{m}^{(1)},\ldots,\mathbf{m}^{(k)} \in \mathbb{R}^{n}$
%
%
%* __Repeat Until Stopping Condition is Met:__  
%
%    * __Update Cluster Assignments__: assign each data to the nearest cluster: 
%    
%    for each data point $i=1,\ldots,m$, set  
%    
%    \begin{equation*}
%    y^{(i)} = \underset{c' \in \{1,\ldots,k\}}{\operatorname{argmin}} \|\mathbf{x}^{(i)} - \mathbf{m}^{(c')}\|^2 , 
%    \tag{1}
%    \end{equation*}
%    
%    * __Update Cluster Means__: determine cluster means for new cluster assignments 
%    
%    for each cluster $c=1,\ldots,k$, set 
%    \begin{equation*}
%    \mathbf{m}^{(c)} = \frac{1}{\mid\{i: y^{(i)}= c\}\mid}{\sum_{i: y^{(i)}= c}\mathbf{x}^{(i)}}     
%    \label{mean}
%    \tag{2}
%    \end{equation*}
%    where $\{i: y^{(i)}= c\}$ represents the set of datapoints belonging to cluster c and $\mid\{i: y^{(i)}= c\}\mid$ the size of cluster c.  

\subsection{Minimizing Clustering Error}
\label{sec_min_k_means}

Algorithm \ref{alg_k_means} can be interpreted as a method for minimizing the clustering error
\begin{align}
\label{equ_clustering_error}
\mathcal{E} ( \{\mathbf{m}^{(\clusteridx)}\}_{\clusteridx=1}^{\nrclusters},\{\clusterassgt^{(\sampleidx)}\}_{\sampleidx=1}^{\samplesize} \mid \{\mathbf{x}^{(\sampleidx)}\}_{\sampleidx=1}^{\samplesize} )& \notag \\ = (1/\samplesize) \sum_{\samplesize=1}^{\samplesize} {\left\|\mathbf{x}^{(\sampleidx)}-\mathbf{m}^{(\clusterassgt^{(\sampleidx)})}\right\|^2}. 
\end{align}
For convenience we will drop the explicit dependence of the clustering error \eqref{equ_clustering_error} 
on the cluster means $\{\mathbf{m}^{(\clusteridx)}\}_{\clusteridx=1}^{\nrclusters}$
and assignments $\{\clusterassgt^{(\sampleidx)}\}_{\sampleidx=1}^{\samplesize}$. 
	
The means update \eqref{equ_cluster_mean_update} within Algorithm \ref{alg_k_means} amounts to minimizing $\mathcal{E}$ 
by varying the cluster means $\mathbf{m}^{(\clusteridx)}$ for fixed assignments $\clusterassgt^{(\sampleidx)}$. 
The assignment update \eqref{equ_cluster_assign_update} within Algorithm \ref{alg_k_means} amounts to minimizing $\mathcal{E}$ 
by varying the assignments $\clusterassgt^{(\sampleidx)}$ with the cluster means $\mathbf{m}^{(\clusteridx)}$ held fixed. Thus, the sequence of cluster means and assignments generated by Algorithm \ref{alg_k_means} 
decreases the clustering error \eqref{equ_clustering_error}. 

Interpreting Algorithm \ref{alg_k_means} as an iterative method for minimizing \eqref{equ_clustering_error} 
provides a natural choice for the stopping criterion in Algorithm \ref{alg_k_means}. We could specify a threshold 
for the required (relative) decrease of the clustering error \eqref{equ_clustering_error} achieved by one additional 
iteration of Algorithm \ref{alg_k_means}. If the decrease of the clustering error \eqref{equ_clustering_error} is smaller 
than this threshold, the iterations in Algorithm \ref{alg_k_means} are stopped. 

Since the clustering error \eqref{equ_clustering_error} is a non-convex function of the cluster means and 
assignments, Algorithm \ref{alg_k_means} might get trapped in a local minimum. It is therefore useful to 
repeat Algorithm \ref{alg_k_means} several times with different initializations for the cluster means and then 
choose the cluster assignment resulting in the smallest empirical risk among all repetitions. 

\section{Soft Clustering}
\label{sec_soft_clustering}

\begin{figure}[htbp]
%\begin{center}
%\begin{tikzpicture}[auto,scale=0.8]
%\draw [thick] (0,2) rectangle ++(0.1cm,0.1cm) node[anchor=east] {\hspace*{0mm}$\vx^{(3)}$};
%%\draw (0,.5) node[cross,red] {};
%\draw [thick] (0.55,2.05)  node[cross,red]  (m1) {}; %{\hspace*{0mm}$\vm^{(1)}$};
%\node  [above=0.5mm of m1]  {\hspace*{1mm} $\vm^{(1)}$};
%\draw [thick] (1,2) rectangle ++(0.1cm,0.1cm) node[anchor=north] {\hspace*{2mm}$\vx^{(2)}$};
%\draw [thick] (2,2) rectangle ++(0.1cm,0.1cm) node[anchor=west,above] {\hspace*{0mm}$\vx^{(1)}$};
%\draw [thick] (8,2) rectangle ++(0.1cm,0.1cm) node[anchor=west,above] {\hspace*{0mm}$\vx^{(5)}$};
%\draw [thick] (5,2) rectangle ++(0.1cm,0.1cm) node[anchor=west,above] {\hspace*{0mm}$\vx^{(6)}$};
%%\draw [thick] (2,4) rectangle ++(0.1cm,0.1cm) node[anchor=west,above] {\hspace*{0mm}$\vx^{(7)}$};
%%\draw[->] (-0.5,0) -- (6.5,0) node[right] {$x_{\rm g}$};
%%\draw[->] (0,-0.5) -- (0,6.5) node[above] {$x_{\rm r}$};
%%\foreach \y/\ytext in { 1/1,2/2,3/3,4/4,5/5} \draw[shift={(0,\y)}] (2pt,0pt) -- (-2pt,0pt) node[left] {$\ytext/5$};  
%%\foreach \x/\xtext in{ 1/1,2/2,3/3,4/4,5/5}\draw[shift={(\x,0)}] (0pt,2pt) -- (0pt,-2pt) node[below] {$\xtext/5$};  
%\end{tikzpicture}
%\end{center}
\begin{center}
\includegraphics[height=5cm]{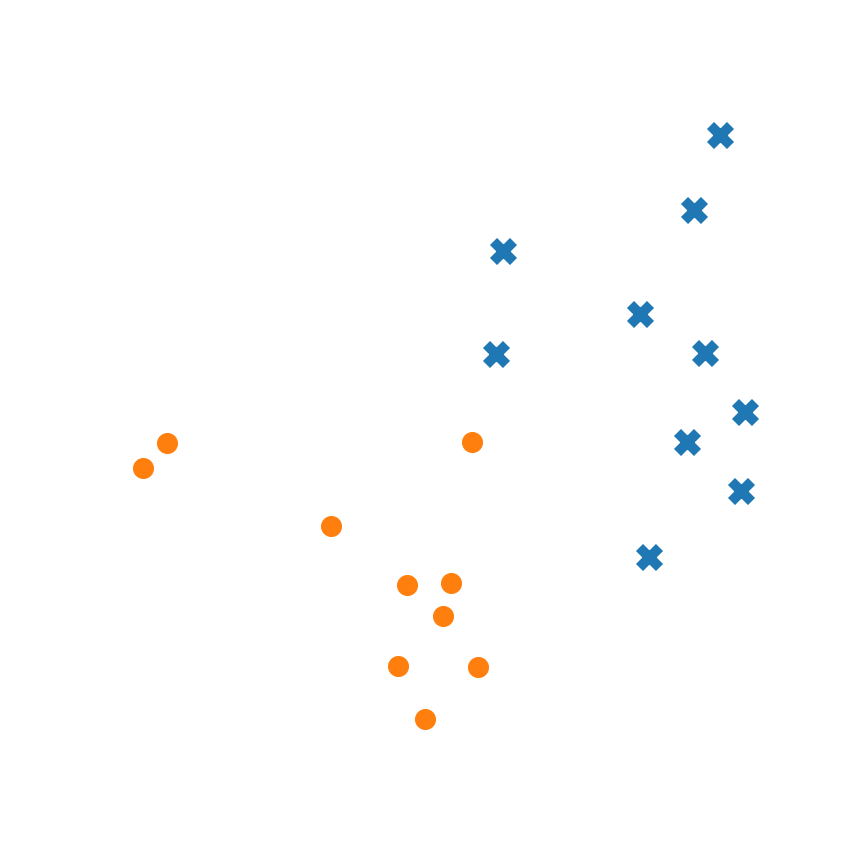}
\vspace*{-10mm}
\end{center}
\caption{Hard clustering of data points. Some data points are closer to the boundary between clusters than others. 
} 
\label{fig_edge_case_hard_clustering}
\end{figure}

The cluster assignments obtained from hard clustering, such as Algorithm \ref{alg:kmeansimpl}, provide 
rather coarse-grained information. Even if two data points $\vx^{(\sampleidx)}, \vx^{(j)}$ are assigned to the 
same cluster $\clusteridx$, their distances to the cluster mean $\vm^{(\clusteridx)}$ might be very different (see Figure \ref{fig_edge_case_hard_clustering}). 
For some applications, we would like to have a more fine-grained information about the cluster assignments. 

%In particular, we are interested in the degree by which a data point belongs to a particular cluster. 
Soft-clustering methods provide such fine-grained information by explicitly modelling the degree (or confidence) by which 
a particular data point belongs to a particular cluster. More precisely, soft-clustering methods track for each data 
point $\vx^{(\sampleidx)}$ the degree of belonging to each of the clusters $\clusteridx \in \{1,\ldots,\nrcluster\}$.  

\subsection{Gaussian Mixture Models} 

A principled approach to defining a degree of belonging to different clusters is to interpret each  
data point $\vx^{(\sampleidx)}$, for $\sampleidx=1,\ldots,\samplesize$, as the realization of a 
random vector. The probability distribution the (random) data point $\vx^{(\sampleidx)}$ depends 
on the cluster to which this data point belongs. Formally, we represent each cluster $\cluster^{(\clusteridx)}$ 
with a \emph{multivariate normal distribution} \cite{Lapidoth09}
\begin{align}
\label{equ_def_mvn_distribution}
 \mathcal{N} (\vx ; {\bm \mu}^{(\clusteridx)}, {\bm \Sigma}^{(\clusteridx)}) & = \nonumber \\ 
 & \hspace*{-20mm}  \frac{1}{\sqrt{{\rm det } \{ 2 \pi  {\bm \Sigma} \} }} \exp\big( - (1/2) (\vx\!-\!{\bm \mu})^{T}  {\bm \Sigma}^{-1} (\vx\!-\!{\bm \mu})  \big)
\end{align} 
of a Gaussian random vector with mean ${\bm \mu}^{(\clusteridx)}$ and (invertible) covariance matrix ${\bf \Sigma}^{(\clusteridx)}$. 

While $k$-means represents each cluster $\cluster^{(\clusteridx)}$ by the cluster mean $\vm^{(\clusteridx)}$,\footnote{Given the cluster means, the cluster assignments can be 
found by identifying the nearest cluster means.} here we represent each cluster by a probability distribution \eqref{equ_def_mvn_distribution}. However, 
since we restrict ourselves to multivariate normal (or Gaussian) distributions, we can parametrize 
the cluster-specific distributions \eqref{equ_def_mvn_distribution} using a mean vector and a covariance 
matrix. In other words we only need to store a vector and a matrix to represent the entire probability 
distribution \eqref{equ_def_mvn_distribution}. 

The index $\clusteridx^{(\sampleidx)}$ of the cluster from which data point $\vx^{(\sampleidx)}$ is 
generated is unknown. Therefore, we model the cluster index $\clusteridx^{(\sampleidx)}$ as a random 
variable with the probability distribution  
\begin{equation} 
\label{equ_def_cluster_prob}
p_{\clusteridx} \defeq \prob{c^{(\sampleidx)}=\clusteridx} \mbox{ for } \clusteridx=1,\ldots,\nrcluster.
\end{equation}  
The (prior) probabilities $p_{\clusteridx}$ are unknown and have to be estimated from observing the data points $\vx^{(\sampleidx)}$. 
If we knew the cluster index $\clusteridx^{(\sampleidx)}=\clusteridx$ of the $\sampleidx$th data point, the distribution of the vector $\vx^{(\sampleidx)}$ 
would be  
\begin{equation}
\label{equ_cond_dist_GMM}
 \prob{\vx^{(\sampleidx)} | \clusteridx^{(\sampleidx)}=\clusteridx } = \mathcal{N} \big(\vx; {\bm \mu}^{(\clusteridx)}, {\bm \Sigma}^{(\clusteridx)} \big)
\end{equation} 
with mean vector ${\bm \mu}^{(\clusteridx)}$ and covariance matrix ${\bm \Sigma}^{(c)}$. 

\begin{figure}
\begin{center}
\begin{tikzpicture}[scale=0.3]
\draw [thick] \boundellipse{0,0}{10}{5} node[right]  {${\bm \mu^{(1)}}$};
 \fill (0,0) circle (2pt) ; 
  \node [right] at (0,5.5) {$ {\bm \Sigma}^{(1)}$} ; 
\draw [thick] \boundellipse{11,1}{-2}{4} node[anchor=south]  {${\bm \mu^{(2)}}$};
 \fill (11,1) circle (2pt) ; 
   \node [right] at (11,5.5) {$ {\bm \Sigma}^{(2)}$} ; 
\draw [thick] \boundellipse{-9,4}{2}{3} node[left,xshift=3mm,yshift=3mm]  {$\,\,{\bm \mu^{(3)}}$}; 
 \fill (-9,4) circle (2pt) ; 
 \node [right] at (-9,8) {$ {\bm \Sigma}^{(3)}$} ; 
\end{tikzpicture}
\end{center}
\caption{The GMM \eqref{equ_def_GMM} yields a probability density function which is a weighted sum of multivariate 
normal distributions $\mathcal{N}({\bm \mu}^{(c)}, {\bm \Sigma}^{(c)})$. The weight of the $c$-th component is 
the cluster probability $ \prob{c^{(\sampleidx)}=c}$.}
\label{fig_GMM_elippses}
\end{figure}
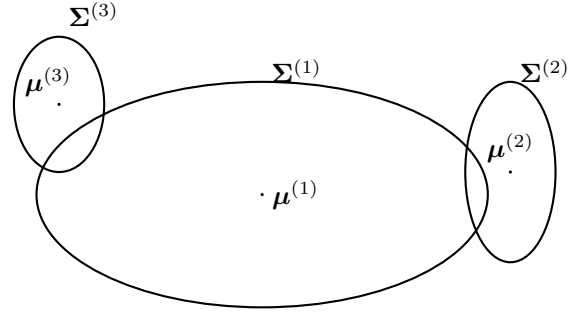

The overall probabilistic model \eqref{equ_cond_dist_GMM}, \eqref{equ_def_cluster_prob} amounts to a {\bf Gaussian mixture model} 
(GMM). Indeed, using the law of total probability, the marginal distribution $\prob{\vx^{(\sampleidx)}}$ (which is the same for all data 
points $\vx^{(\sampleidx)}$) is a \emph{mixture of Gaussians},  
\begin{equation} 
\label{equ_def_GMM}
 \prob{\vx^{(\sampleidx)}} = \sum_{\clusteridx=1}^{\nrclusters} \underbrace{\prob{\clusterassgt^{(\sampleidx)}=\clusteridx}}_{p_{\clusteridx}}  \underbrace{\prob{\vx^{(\sampleidx)} | \clusterassgt^{(\sampleidx)}=\clusteridx}}_{\mathcal{N}(\vx^{(\sampleidx)};{\bm \mu}^{(\clusteridx)}, {\bm \Sigma}^{(\clusteridx)})}. 
\end{equation} 
Figure \ref{fig_GMM_elippses} depicts a GMM using $\nrclusters=3$ clusters. 

It is important to note that the cluster assignments $\clusterassgt^{(\sampleidx)}$ are 
hidden (unobserved) random variables. We thus have to infer or estimate these variables
 from the observed data points $\mathbf{x}^{(\sampleidx)}$ which are i.i.d.\ realizations of the GMM \eqref{equ_def_GMM}.

Modelling the cluster assignments $\clusteridx^{(\sampleidx)}$ as (realizations of) random variables 
lends naturally to a precise definition for the degree of belonging to a particular cluster.  
In particular, we define the degree $y^{(\sampleidx)}_{\clusteridx}$ of data point $\vx^{(\sampleidx)}$ belonging 
to the cluster $\cluster^{(\clusteridx)}$ as the ``a-posteriori'' probability of the assignment variable $\clusteridx^{(\sampleidx)}$ being equal to $\clusteridx \in \{1,\ldots,\nrcluster\}$, 
\begin{equation}
\label{equ_def_deg_belonging_prob}
y^{(\sampleidx)}_{\clusteridx} \defeq \prob{ \clusteridx^{(\sampleidx)} = \clusteridx | \dataset}.
\end{equation} 
For each data point $\vx^{(\sampleidx)}$, the degree of belonging \eqref{equ_def_deg_belonging_prob} 
involves a conditioning on the entire dataset $\dataset$. This is reasonable since the degree of belonging 
for each data point typically depends on the overall (cluster) geometry of the entire dataset $\dataset$.

The degrees of belonging $y^{(\sampleidx)}_{\clusteridx}$ sum to one, 
\begin{equation} 
\label{equ_dob_sum_to_one}
\sum_{\clusteridx=1}^{\nrcluster} y^{(\sampleidx)}_{\clusteridx}=1 \mbox{ for each } \sampleidx=1,\ldots,\samplesize.
\end{equation}  

%A probabilistic model for the observed data points $\vx^{(\sampleidx)}$ is obtained by considering each data point $\vx^{(\sampleidx)}$ being the result of a random draw from the 
%distribution $\mathcal{N} (\vx ; {\bm \mu}^{(c^{(\sampleidx)})}, {\bm \Sigma}^{(c^{(\sampleidx)})})$ with some cluster $c^{(\sampleidx)}$. 
%Since the cluster indices $c^{(\sampleidx)}$ are unknown,\footnote{After all, the goal of soft-clustering is to estimate the cluster indices $c^{(\sampleidx)}$.} 
%we model them as random variables. In particular, we model the cluster indices $c^{(\sampleidx)}$ as i.i.d. with probabilities 

%\end{document}

\subsection{Expectation Maximization} 

Using the GMM \eqref{equ_def_GMM} for explaining the observed data points $\vx^{(\sampleidx)}$ 
turns the clustering problem into a {\bf statistical inference} or {\bf parameter estimation problem} \cite{kay,LC}. 
We have to estimate (approximate) the true underlying cluster probabilities $p_{\clusteridx}$ (see \eqref{equ_def_cluster_prob}), 
cluster means ${\bm \mu}^{(\clusteridx)}$ and cluster covariance matrices ${\bm \Sigma}^{(\clusteridx)}$ 
(see \eqref{equ_cond_dist_GMM}) from the observed data points $\dataset = \{ \vx^{(\sampleidx)} \}_{\sampleidx=1}^{\samplesize}$, 
which are drawn from the probability distribution \eqref{equ_def_GMM}. 
%Having estimates for the GMM parameters allows us to compute an estimate (approximation) of the ``a-posteriori'' probability 
%$\prob\{ c^{(\sampleidx)}= c | \dataset \}$, which we consider as the ``true'' degree of data point $\vx^{(\sampleidx)}$ belonging to cluster $c$. 

We denote the estimates for the GMM parameters by $\hat{p}_{\clusteridx} (\approx p_{\clusteridx})$, $\vm^{(\clusteridx)} (\approx {\bm \mu}^{(\clusteridx)})$ and 
$\mathbf{C}^{(\clusteridx)} (\approx {\bm \Sigma}^{(\clusteridx)})$, respectively. Based on these estimates, we can then compute an estimate $\hat{y}_{\clusteridx}^{(\sampleidx)}$ 
of the (a-posterior) probability 
\begin{equation} 
y_{\clusteridx}^{(\sampleidx)} = \prob{c^{(\sampleidx)} = \clusteridx \mid \dataset }
\end{equation} 
of the $\sampleidx$-th data point $\vx^{(\sampleidx)}$ belonging to cluster $\clusteridx$, given the observed dataset $\dataset$. 

It turns out that this estimation problem becomes significantly easier by operating in an alternating 
fashion: In each iteration, we first compute a new estimate $\hat{p}_{c}$ of the cluster probabilities 
$p_{\clusteridx}$, given the current estimate $\vm^{(\clusteridx)}, \mathbf{C}^{(\clusteridx)}$ for the 
cluster means and covariance matrices. Then, using this new estimate $\hat{p}_{\clusteridx}$ for the 
cluster probabilities, we update the estimates $\vm^{(\clusteridx)}, \mathbf{C}^{(\clusteridx)}$ of the 
cluster means and covariance matrices. Then, using the new estimates $\vm^{(\clusteridx)}, \mathbf{C}^{(\clusteridx)}$, 
we compute a new estimate $\hat{p}_{\clusteridx}$ and so on. By repeating these two update steps, 
we obtain an iterative soft-clustering method which is summarized in Algorithm \ref{alg:softclustering}. 

%Let us, for ease of exposition, restrict to the case of $k\!=\!2$ clusters $\clusteridx\!\in\!\{1,2\}$. The generalization to $k\!>\!2$ clusters is easy. 
%For $k=2$, we only need to track the degree of belonging $y_{1}^{(\sampleidx)}$ to cluster $\clusteridx=1$ since $y_{2}^{(\sampleidx)}\!=\!1\!-\!y_{1}^{(\sampleidx)}$ (see \eqref{equ_dob_sum_to_one}). 
%The estimates of the probabilities can be written as $\hat{p}_{1} = \samplesize_{1}/\samplesize$ and $\hat{p}_{2} = \samplesize_{2}/\samplesize$ 
%with $\samplesize_{1} = \sum_{\sampleidx=1}^{\samplesize} y^{(\sampleidx)}_{1}$ and $\samplesize_{1} = \sum_{\sampleidx=1}^{\samplesize} y^{(\sampleidx)}_{2}$. 
%The quantities $\samplesize_{1}, \samplesize_{2}$ can be interpreted as {\bf effective cluster sizes}. 
\begin{algorithm}[htbp]
\caption{``A Soft-Clustering Algorithm'' \cite{BishopBook}}\label{alg:softclustering}
\begin{algorithmic}[1]
\renewcommand{\algorithmicrequire}{\textbf{Input:}}
\renewcommand{\algorithmicensure}{\textbf{Output:}}
\Require   dataset $\dataset=\{ \vx^{(\sampleidx)}\}_{\sampleidx=1}^{\samplesize}$; number $k$ of clusters. 
\Statex\hspace{-6mm}{\bf Initialize:} use initial guess for GMM parameters $\{\mathbf{m}^{(\clusteridx)},\mathbf{C}^{(\clusteridx)},\hat{p}_{\clusteridx}\}_{\clusteridx=1}^{\nrcluster}$ 
\Repeat
\vspace*{2mm}
\State for each data point $\vx^{(\sampleidx)}$ and cluster $\clusteridx \in \{1,\ldots,\nrcluster\}$, update degrees of belonging
\vspace*{-1mm}
\begin{align}
\label{equ_update_soft_cluster_assignm}
y_{c}^{(\sampleidx)} & =  \frac{\hat{p}_{\clusteridx} \mathcal{N}(\vx^{(\sampleidx)};\mathbf{m}^{(\clusteridx)},\mathbf{C}^{(\clusteridx)})}{\sum_{\clusteridx'=1}^{\nrcluster} \hat{p}_{\clusteridx'}\mathcal{N}(\vx^{(\sampleidx)};\mathbf{m}^{(\clusteridx')},\mathbf{C}^{(\clusteridx')})} %\nonumber \\
%y_{2}^{(\sampleidx)} & = 1- y_{1}^{(\sampleidx)} 
\end{align}
\State for each cluster $\clusteridx \in \{1,\ldots,\nrcluster\}$, update estimates of 
\begin{itemize} 
\item cluster probability $\hat{p}_{\clusteridx}\!=\!\samplesize_{\clusteridx}/\samplesize$, with effective cluster size $\samplesize_{\clusteridx}\!=\!\sum\limits_{\sampleidx=1}^{\samplesize} y_{\clusteridx}^{(\sampleidx)}$
\item compute cluster mean (estimate) 
$$\hspace*{-10mm}\mathbf{m}^{(\clusteridx)} = (1/\samplesize_{\clusteridx}) \sum\limits_{\sampleidx=1}^{\samplesize} y_{\clusteridx}^{(\sampleidx)} \vx^{(\sampleidx)}$$
\item compute cluster covariance matrix (estimate) 
$$\hspace*{-10mm}\mathbf{C}^{(\clusteridx)}  = (1/\samplesize_{\clusteridx}) {\sum\limits_{\sampleidx=1}^{\samplesize} y_{\clusteridx}^{(\sampleidx)} \big(\vx^{(\sampleidx)}\!-\!\mathbf{m}^{(\clusteridx)}\big)   \big(\vx^{(\sampleidx)}\!-\!\mathbf{m}^{(\clusteridx)}\big)^{T} }$$
\end{itemize}
\vspace*{1mm}
\Until convergence
\Ensure soft cluster assignments $\vy^{(\sampleidx)}=(y_{1}^{(\sampleidx)},\ldots,y_{k}^{(\sampleidx)})^{T}$ for each data point $\vx^{(\sampleidx)}$ 
\end{algorithmic}
\end{algorithm}

As with hard clustering (see Section \ref{sec_min_k_means}), we can also interpret soft clustering as an 
optimization problem. In particular, Algorithm \ref{alg:softclustering} aims at choosing GMM parameter estimates 
$\{ \vm^{(\clusteridx)}, \mathbf{C}^{(\clusteridx)}, \hat{p}_{\clusteridx} \}_{\clusteridx=1}^{\nrcluster}$ that minimize the \emph{soft clustering error} 
\begin{align} 
\label{equ_def_emp_risk_soft_clustering}
\emperror^{(\rm SC)} % \big( \{ \vm^{(\clusteridx)}, \mathbf{C}^{(\clusteridx)}, \hat{p}_{\clusteridx} \}_{\clusteridx=1}^{\nrcluster} \mid \dataset \big)
 \defeq 
 - \log \prob{\dataset; \{ \vm^{(\clusteridx)}, \mathbf{C}^{(\clusteridx)}, \hat{p}_{\clusteridx} \}_{\clusteridx=1}^{\nrcluster}}. 
\end{align} 
The interpretation of Algorithm \ref{alg:softclustering} as a method for minimizing the empirical 
risk \eqref{equ_def_emp_risk_soft_clustering} suggests to monitor the decrease of the empirical 
risk $\emperror^{(\rm SC)}$ 
to decide when to stop iterating (``convergence test''). % the termination criterion. % Algorithm \ref{alg:softclustering}. 

Similar to $k$-means Algorithm \ref{alg_k_means}, also the soft clustering Algorithm \ref{alg:softclustering} 
suffers from the problem of getting stuck in local minima of the empirical risk \eqref{equ_def_emp_risk_soft_clustering}. 
Therefore, as for $k$-means Algorithm \ref{alg_k_means} (see Section \ref{sec_hard_clustering}), one typically runs 
Algorithm \ref{alg:softclustering} several times. Each run uses a different choice for the initial GMM parameters in 
Algorithm \ref{alg:softclustering} and results in different estimates $ \{ \vm^{(c)}, \mathbf{C}^{(c)}, \hat{p}_{c} \}_{c=1}^{k}$ for 
the GMM parameters. We then pick the GMM parameters obtained in the run achieving the smallest empirical risk \eqref{equ_def_emp_risk_soft_clustering}.  

We note that the empirical risk \eqref{equ_def_emp_risk_soft_clustering} underlying the 
soft-clustering Algorithm \ref{alg:softclustering} is essentially a {\bf log-likelihood function}. 
Thus, Algorithm \ref{alg:softclustering} can be interpreted as an {\bf approximate maximum 
likelihood} estimator for the true underlying GMM parameters $\{{\bm \mu}^{(c)},{\bm \Sigma}^{(c)},p_{c}\}_{c=1}^{k}$. 
In particular, Algorithm \ref{alg:softclustering} is an instance of a generic approximate maximum likelihood 
technique referred to as {\bf expectation maximization} (EM) (see \cite[Chap. 8.5]{hastie01statisticallearning} for more details). 
The interpretation of Algorithm \ref{alg:softclustering} as a special case of EM allows to characterize the 
behaviour of Algorithm \ref{alg:softclustering} using existing convergence results for EM methods \cite{XuJordan1996}. 
%???? Discuss EM principle in more detail; point out relation to recent work of Arora on Factorization methods such as Dictionary Learning, proving global convergence 
%New Algorithms for Learning Incoherent and Overcomplete Dictionaries, Sanjeev Arora, Rong Ge, Ankur Moitra
%https://arxiv.org/pdf/1510.06096.pdf
%
%Arora, Sanjeev, Rong Ge, Ankur Moitra, and Sushant Sachdeva.
%?Provable ICA with Unknown Gaussian Noise, and Implications
%for Gaussian Mixtures and Autoencoders.? Algorithmica 72, no. 1
%(March 4, 2015): 215?236.
%??????

\subsection{GMM and $k$-means}

We finally note that $k$-means hard clustering can be interpreted as an extreme case of 
soft-clustering Algorithm \ref{alg:softclustering}. Indeed, consider fixing the cluster 
covariance matrices $ {\bm \Sigma}^{(c)}$ within the GMM \eqref{equ_cond_dist_GMM} 
to be the scaled identity: 
\begin{equation}
\label{equ_def_special_case}
 {\bm \Sigma}^{(\clusteridx)}= \sigma^{2} \mathbf{I} \mbox{ for all } \clusteridx \in \{1,\ldots,\nrclusters\}.  
\end{equation} 
Here, we assume the covariance matrix \eqref{equ_def_special_case}, with a particular value 
for $\sigma^{2}$, to be the actual ``correct'' covariance matrix for cluster $c$. The estimates 
$\mathbf{C}^{(\clusteridx)}$ for the covariance matrices are then trivially given by $\mathbf{C}^{(\clusteridx)} =  {\bm \Sigma}^{(\clusteridx)}$, 
i.e., we can omit the covariance matrix updates in Algorithm \ref{alg:softclustering}. Moreover, when 
choosing a very small variance $\sigma^{2}$ in \eqref{equ_def_special_case}), the update 
\eqref{equ_update_soft_cluster_assignm} tends to enforce $y_{c}^{(\sampleidx)} \in \{0,1\}$, i.e., 
each data point $\vx^{(\sampleidx)}$ is associated to exactly one cluster $c$, whose cluster mean 
$\vm^{(c)}$ is closest to the data point $\vx^{(\sampleidx)}$. Thus, for $\sigma^{2} \rightarrow 0$, 
the soft-clustering update \eqref{equ_update_soft_cluster_assignm} reduces to the hard cluster 
assignment update \eqref{equ_cluster_assign_update} of the $k$-means Algorithm \ref{alg_k_means}.

\section{Density Based Clustering} 
\label{sec_density_clustering}

Both, $k$-means and GMM, cluster data points according to their mutual Euclidean distances. These two methods 
judge if they data points $\vx^{(i)}$ and $\vx^{(j)}$ are similar solely based on their Euclidean distance $\|\vx^{(i)} - \vx^{(j)} \|$. 
Here, two data points cannot be similar if their Euclidean distance is large. 

Some applications involve data points which are organized according to a non-Euclidean structure. Consider an 
image database which contains images of animals. If we want to organize (cluster) those images according to 
the species of the depicted animal, it might not be useful to measure the similarity of images based on the Euclidean 
distance between the corresponding vectors of red, green and blue pixel intensities. 

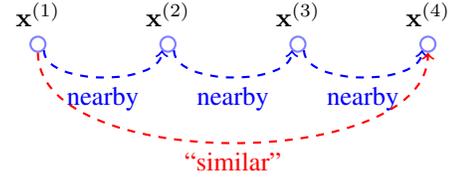
\begin{figure}[htbp]
\begin{center}
\tikzstyle{place}=[circle,draw=blue!50,thick,
                   inner sep=0pt,minimum size=2mm]
\tikzstyle{transition}=[rectangle,draw=black!50,fill=black!20,thick,
                        inner sep=0pt,minimum size=4mm]
\begin{tikzpicture}[auto,yscale=0.8,xscale=0.6]
 \node[place] at (1,3) (x1) {} ; 
  \node[place,right = 1.5cm of x1] (x2){} ; 
  \node[place,right = 1.5cm of x2] (x3) {}; 
    \node[place,right = 1.5cm of x3] (x4) {}; 
 % \node[transition] (leave critical) [right of=critical] {};
%  \node[transition] (enter critical) [left of=critical]  {};
  \node [above] at (x2.north) {$\mathbf{x}^{(2)}$};
  \node [above] at (x1.north) {$\mathbf{x}^{(1)}$};
   \node [above] at (x3.north) {$\mathbf{x}^{(3)}$};
\node [above] at (x4.north) {$\mathbf{x}^{(4)}$};
\draw[thick,blue,->,dashed] (x1.south east) .. controls +(down:0.6cm) and +(down:0.6cm) .. node [below] {nearby} (x2.south west);
\draw[thick,blue,->,dashed] (x2.south east) .. controls +(down:0.6cm) and +(down:0.6cm) .. node [below] {nearby} (x3.south west);
\draw[thick,blue,->,dashed] (x3.south east) .. controls +(down:0.6cm) and +(down:0.6cm) .. node [below] {nearby} (x4.south west);
\draw[thick,red,->,dashed] (x1.south) .. controls +(down:2cm) and +(down:2cm) .. node [below] {``similar''} (x4.south);
  
%  \draw[->] (-0.5,0) -- (10.5,0) node[right] {$x_{\rm g}$};
%\draw[->] (0,-0.5) -- (0,6.5) node[above] {$x_{\rm r}$};
%\foreach \y/\ytext in { 1/1,2/2,3/3,4/4,5/5} \draw[shift={(0,\y)}] (2pt,0pt) -- (-2pt,0pt) node[left] {$\ytext/5$};  
%\foreach \x/\xtext in{ 1/1,2/2,3/3,4/4,5/5}\draw[shift={(\x,0)}] (0pt,2pt) -- (0pt,-2pt) node[below] {$\xtext/5$};  
\end{tikzpicture}
\end{center}
\caption{Two data points $\mathbf{x}^{(1)}$ and $\mathbf{x}^{(4)}$ are considered ``similar'' if they can be reached
 by a sequence of nearby intermediate data points. 
} 
\label{fig_DBSCAN}
\end{figure}

One example for a non-Euclidean structure is based on the notion of connectivity. Two data points are considered 
similar if they can be reached  (if they are connected) by a sequence of nearby data points (see Figure~\ref{fig_DBSCAN} ). 
Here, two data points can be similar even if their Euclidean distance is large. 

Density-based spatial clustering of applications with noise (DBSCAN) is one example for a hard clustering method 
using connectivity as similarity measure \cite{DBSCAN}. DBSCAN requires two parameters: the maximum Euclidean 
distance $\varepsilon$ between two data points to be considered as nearby. 
The second input parameter required by DBSCAN is the minimum number of nearby data points $m_{\rm near}$ 
required to consider a data poit to be a \emph{core data point}. Each cluster obtained from DBSCAN is constituted 
by some core points and all its nearby (within Euclidean distance $\varepsilon$) data points. 

In contrast to $k$-Means and GMM, DBSCAN does not require the specification of the number of clusters. 
Instead, the number of clusters is determined automatically based on the intrinsic structure of the dataset. 
Moreover, DBSCAN allows to detect outliers which can be interpreted as degenerated clusters 
consisting of only one single data point. 

\begin{figure}[htbp]
\begin{center}
\includegraphics[width=1\linewidth]{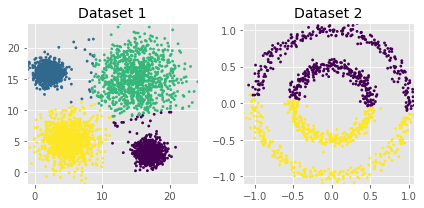}
\end{center}
\caption{Applying $k$-means algorithm to the two sample datasets.}
\label{fig_DBSCAN2}
\end{figure}

Figure~\ref{fig_DBSCAN2} illustrates a dataset $\mathbb{X}^{(1)}$ with Euclidean cluster structure 
and another dataset $\mathbb{X}^{(2)}$ with a non-Euclidean cluster structure. The colouring of data 
points in each dataset represents the clustering obtained from $k$-means using $k=4$ for $\mathbb{X}^{(1)}$ 
and $k=2$ for $\mathbb{X}^{(2)}$. While $k$-means produces a reasonable clustering of the first dataset 
$\mathbb{X}^{(1)}$, it fails to find the intrinsic cluster structure of the second dataset $\mathbb{X}^{(2)}$.

\begin{figure}[htbp]
\begin{center}
\includegraphics[width=1\linewidth]{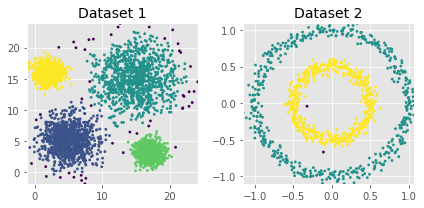}
\end{center}
\caption{Applying DBSCAN algorithm to the two sample datasets.}
\label{fig_DBSCAN3}
\end{figure}

Figure~\ref{fig_DBSCAN3} again shows the datasets $\mathbb{X}^{(1)}$ and $\mathbb{X}^{(2)}$ but with data points 
clustered using DBSCAN instead of $k$-means. It can be seen that DBSCAN is able to produce reasonable clustering 
on both datasets and also identifies potential outliers.

\section*{Acknowledgments}
We thank the students of the Aalto courses ``Machine Learning: Basic Principles'', ``Artificial Intelligence'' 
and ``Machine Learning with Python'' for their constructive and critical feedback. This feedback was instrumental 
for the authors to learn how to teach clustering. 

\bibliographystyle{IEEEtran}
\bibliography{/Users/alexanderjung/Literature}

\end{document}